\documentclass[pdflatex,sn-mathphys-num]{sn-jnl}

\usepackage{graphicx}%
\usepackage{multirow}%
\usepackage{amsmath,amssymb,amsfonts}%
\usepackage{amsthm}%
\usepackage{mathrsfs}%
\usepackage[title]{appendix}%
\usepackage{xcolor}%
\usepackage{textcomp}%
\usepackage{manyfoot}%
\usepackage{booktabs}%
\usepackage{caption}
\usepackage{algorithm}%
\usepackage{algorithmicx}%
\usepackage{algpseudocode}%
\usepackage{listings}%

\usepackage{footmisc}
\usepackage{graphicx}
\usepackage{amsmath}
\usepackage{bm}
\usepackage{xcolor}
\usepackage{pifont}
\usepackage{adjustbox} 
\usepackage{booktabs}
\usepackage{threeparttable}
\usepackage{multirow}
\usepackage{soul}
\usepackage{array}  
\newcolumntype{P}[1]{>{\centering\arraybackslash}p{#1}}
\usepackage{algorithm}
\usepackage{algpseudocode}
\usepackage[T1]{fontenc}
\usepackage[utf8]{inputenc}

\newcommand{\method}{{\fontfamily{ppl}\selectfont
ePAI}}

\newcommand{\internalauc}{0.985}
\newcommand{\internalaucrange}{0.974--0.994}
\newcommand{\internalallsizesens}{97.1\%}
\newcommand{\internalallsizesensrange}{94.4--98.5\%}
\newcommand{\internalsmallsens}{95.3\%}
\newcommand{\internalsmallsensrange}{84.5--98.7\%}
\newcommand{\internalsmallaucrange}{0.939--0.999}
\newcommand{\internalsmallpatients}{43}
\newcommand{\internallargesens}{97.5\%}
\newcommand{\internallargesensrange}{94.6--98.8\%}
\newcommand{\internallargepatients}{236}
\newcommand{\internalspec}{98.7\%}
\newcommand{\internalspecrange}{96.6--99.5\%}
\newcommand{\internalnormalpatients}{303}
\newcommand{\internalallpatients}{1,009}
\newcommand{\internalallsizelocalizationsens}{94.6\%}
\newcommand{\internalallsizelocalizationsensrange}{92.4–96.1\%}
\newcommand{\internalsmalllocalizationsens}{88.2\%}
\newcommand{\internalsmalllocalizationsensrange}{80.0--93.3\%}

\newcommand{\externalauc}{0.971}
\newcommand{\externalaucrange}{0.967--0.976}
\newcommand{\externalallsizesens}{97.0\%}

\newcommand{\externalsmallsens}{91.5\%}
\newcommand{\externalsmallsensrange}{89.0--93.4\%}
\newcommand{\externalsmallaucrange}{0.918--0.945}
\newcommand{\externalsmallpatients}{610}
\newcommand{\externallargesens}{98.3\%}
\newcommand{\externallargesensrange}{97.7--98.7\%}
\newcommand{\externallargepatients}{2,529}
\newcommand{\externalspec}{88.0\%}
\newcommand{\externalspecrange}{87.0--89.0\%}
\newcommand{\externalnormalpatients}{4,019}
\newcommand{\externalpdacpatients}{3,139}
\newcommand{\externalallpatients}{7,158}
\newcommand{\externalcenters}{6}
\newcommand{\externalallsizelocalizationsens}{88.7\%}
\newcommand{\externalallsizelocalizationsensrange}{87.4--89.9\%}
\newcommand{\externalsmalllocalizationsens}{79.7\%}
\newcommand{\externalsmalllocalizationsensrange}{75.3--84.1\%}

\newcommand{\prediagnosticdetected}{75}
\newcommand{\prediagnosticallpatients}{159}
\newcommand{\prediagnosticcenters}{3}
\newcommand{\prediagnostictime}{347} 
\newcommand{\prediagnosticlocalized}{75\%} 
\newcommand{\prediagnosticductdetected}{85\%} 

\newcommand{\numofreaders}{30}

\newcommand{\readerpdacspec}{95.4\%}

\newcommand{\outperformreadersens}{50.3\%}
\newcommand{\outperformreadersensrange}{37.6--62.1\%}

\newcommand{\deltareaderplusepaiprediagnocticsens}{6.9\%}
\newcommand{\deltareaderplusepaidiagnocticsens}{19.6\%}

\theoremstyle{thmstyleone}%
\theoremstyle{thmstyletwo}%

\theoremstyle{thmstylethree}%

\begin{document}

\title[Article Title]{Early and Prediagnostic Detection of Pancreatic Cancer from Computed Tomography}



\author[1]{\fnm{Wenxuan} \sur{Li}} \equalcont{These authors contributed equally to this work.}
\author[1,2,3]{\fnm{Pedro R. A. S.} \sur{Bassi}} \equalcont{These authors contributed equally to this work.}
\author[4]{\fnm{Lizhou} \sur{Wu}} \equalcont{These authors contributed equally to this work.}
\author[1]{\fnm{Xinze} \sur{Zhou}} \equalcont{These authors contributed equally to this work.}
\author[5]{\fnm{Yuxuan} \sur{Zhao}} \equalcont{These authors contributed equally to this work.}
\author[1]{\fnm{Qi} \sur{Chen}} \equalcont{These authors contributed equally to this work.}
\author[6]{\fnm{Szymon} \sur{Płotka}}\equalcont{These authors contributed equally to this work.}
\author[1]{\fnm{Tianyu} \sur{Lin}}
\author[7,8]{\fnm{Zheren} \sur{Zhu}}
\author[7]{\fnm{Marisa} \sur{Martin}}
\author[7]{\fnm{Justin} \sur{Caskey}}
\author[9]{\fnm{Shanshan} \sur{Jiang}}
\author[10]{\fnm{Xiaoxi} \sur{Chen}}
\author[11]{\fnm{Jarosław B.} \sur{Ćwikła}}
\author[12]{\fnm{Artur} \sur{Sankowski}}
\author[13]{\fnm{Yaping} \sur{Wu}}
\author[14]{\fnm{Tom} \sur{Lu}}
\author[14]{\fnm{Ebenezer} \sur{Daniel}}
\author[3]{\fnm{Sergio} \sur{Decherchi}}
\author[2,3,15]{\fnm{Andrea} \sur{Cavalli}}
\author[14]{\fnm{Chandana} \sur{Lall}}
\author[14]{\fnm{Cristian} \sur{Tomasetti}}
\author[13]{\fnm{Yaxing} \sur{Guo}}
\author[13]{\fnm{Xuan} \sur{Yu}}
\author[16]{\fnm{Yuqing} \sur{Cai}}
\author[17]{\fnm{Hualin} \sur{Qiao}}
\author[18]{\fnm{Jie} \sur{Bao}}
\author[18]{\fnm{Chenhan} \sur{Hu}}
\author[18]{\fnm{Ximing} \sur{Wang}}
\author[19,20]{\fnm{Arkadiusz} \sur{Sitek}}
\author[9]{\fnm{Kai} \sur{Ding}}
\author[9]{\fnm{Heng} \sur{Li}}

\author*[13,21]{\fnm{Meiyun} \sur{Wang}}
\author*[5]{\fnm{Dexin} \sur{Yu}}
\author*[4]{\fnm{Guang} \sur{Zhang}}
\author*[7]{\fnm{Yang} \sur{Yang}}
\author*[7]{\fnm{Kang} \sur{Wang}}
\author*[1]{\fnm{Alan L.} \sur{Yuille}}
\author*[1]{\fnm{Zongwei} \sur{Zhou}}\email{zzhou82@jh.edu}

\affil[1]{\orgname{Johns Hopkins University, Baltimore, MD, USA}}
\affil[2]{\orgname{University of Bologna, Bologna, Italy}}
\affil[3]{\orgname{Istituto Italiano di Tecnologia, Genova, Italy}}
\affil[4]{\orgname{Shandong Provincial Qianfoshan Hospital, Shandong, China}}
\affil[5]{\orgname{Qilu Hospital of Shandong University, Shandong, China}}
\affil[6]{\orgname{Jagiellonian University, Krakow, Poland}}
\affil[7]{\orgname{University of California, San Francisco, CA, USA}}
\affil[8]{\orgname{University of California, Berkeley, CA, USA}}
\affil[9]{\orgname{Johns Hopkins Medicine, Baltimore, MD, USA}}
\affil[10]{\orgname{University of Illinois Urbana-Champaign, IL, USA}}
\affil[11]{\orgname{University of Warmia and Mazury, Olsztyn, Poland}}
\affil[12]{\orgname{National Medical Institute of the Ministry of Internal Affairs and Administration, Warsaw, Poland}}
\affil[13]{\orgname{Henan Provincial People’s Hospital \& The People’s Hospital of Zhengzhou University, Zhengzhou, China}}
\affil[14]{\orgname{City of Hope National Medical Center, Duarte, CA, USA}}
\affil[15]{\orgname{Centre Européen de Calcul Atomique et Moléculaire, École Polytechnique Fédérale de Lausanne, Lausanne, Switzerland}}
\affil[16]{\orgname{Northeastern University, Boston, MA, USA}}
\affil[17]{\orgname{Rutgers University, New Brunswick, NJ, USA}}
\affil[18]{\orgname{The First Affiliated Hospital of Soochow University, Soochow, China}}
\affil[19]{\orgname{Massachusetts General Hospital, Boston, MA, USA}}
\affil[20]{\orgname{Harvard University, Cambridge, MA, USA}}
\affil[21]{\orgname{Biomedical Research Institute, Henan Academy of Sciences, Zhengzhou, China}}

\abstract{
Pancreatic ductal adenocarcinoma (PDAC), one of the deadliest solid malignancies, is often detected at a late and inoperable stage. Retrospective reviews of prediagnostic CT scans, when conducted by expert radiologists aware that the patient later developed PDAC, frequently reveal lesions that were previously overlooked. To help detecting these lesions earlier, we developed an automated system named \method\ (\ul{e}arly \ul{P}ancreatic cancer detection with \ul{A}rtificial \ul{I}ntelligence). It was trained on data from 1,598 patients from a single medical center. In the internal test involving \internalallpatients\ patients, \method\ achieved an area under the receiver operating characteristic curve (AUC) of \internalsmallaucrange, a sensitivity of \internalsmallsens, and a specificity of \internalspec\ for detecting small PDAC less than 2 cm in diameter, precisely localizing PDAC as small as 2 mm. In an external test involving \externalallpatients\ patients across \externalcenters\ centers, \method\ achieved an AUC of \externalsmallaucrange, a sensitivity of \externalsmallsens, and a specificity of \externalspec, precisely localizing PDAC as small as 5 mm. Importantly, \method\ detected PDACs on prediagnostic CT scans obtained 3 to 36 months before clinical diagnosis that had originally been overlooked by radiologists. It successfully detected and localized PDACs in \prediagnosticdetected\ of \prediagnosticallpatients\ patients, with a median lead time of \prediagnostictime\ days before clinical diagnosis. Our multi-reader study showed that \method\ significantly outperformed \numofreaders\ radiologists by \outperformreadersens\ (P < 0.05) in sensitivity while maintaining a comparable specificity of \readerpdacspec\ in detecting PDACs early and prediagnostic. These findings suggest its potential of \method\ as an assistive tool to improve early detection of pancreatic cancer.
}

\maketitle

\section{Main}\label{sec:introduction}

Pancreatic ductal adenocarcinoma (PDAC) is one of the deadliest cancers worldwide. Each year, more than 495,000 new cases diagnosed globally, and almost the same number of patients die from the disease, highlighting its exceptionally high fatality rate. Despite advances in cancer treatment, the overall five-year survival rate of PDAC remains around 13\%~\cite{siegel2025cancerstats}.

Survival is low largely because early detection is uncommon. In most patients, PDAC is detected only after it has already spread beyond the pancreas, when curative treatment is no longer possible. In contrast, when PDAC is detected while still localized to the pancreas, five-year survival increases to approximately 40–45\%, compared with about 3\% when detection occurs after distant spread. 

Computed tomography (CT) offers a scalable path because it is already used at massive global volume: about 300 million CT examinations are performed worldwide each year, and roughly 40\% are contrast-enhanced. Even if only a subset are contrast-enhanced abdominal CT, this still corresponds to tens of millions of scans per year worldwide that already include the pancreas. This scale creates a practical opportunity for opportunistic detection of PDAC on scans obtained for other clinical reasons, while also providing spatial localization that blood-based tests and molecular assays cannot. 

However, CT alone has not proven effective for detecting small, early-stage cancers or precancerous lesions. Recognizing subtle early changes is difficult, even for experienced radiologists. Because most abdominal CT scans are performed for non-pancreatic indications, radiologists may not sufficiently focus on the pancreas. When radiologists retrospectively review prediagnostic CT scans, their sensitivity is often low, and agreement between readers is limited. For example, \citet{mukherjee2022gastroenterology} reported sensitivities of only 33.3\% and 31.1\% among two radiologists. These results highlight the limits of visual interpretation using conventional methods.

Several retrospective studies have shown that early signs of PDAC are often present on CT scans months or even years before clinical diagnosis. These signs include pancreatic duct dilatation, pancreatic atrophy, and focal intrapancreatic late enhancement. Across studies, such findings were visible in 16\% to 71\% of prediagnostic CT scans obtained 3 to 36 months before diagnosis~\cite{hoogenboom2021pancreatic,konno2023retrospective,chung2022clinical,toshima2021ct,singh2020computerized}. These observations suggest that prediagnostic CT scans contain valuable information for detecting PDAC at an earlier and potentially more treatable stage.

This gap between the diagnostic potential of prediagnostic CT scans and human performance points to a clear need for computational assistance. Artificial intelligence (AI) is well suited for this role. AI has achieved expert-level performance in many cancer detection tasks~\cite{esteva2017dermatologist,li2024well,bassi2025learning,bassi2025scaling}. However, relatively few studies have focused on early or small PDAC detection on CT scans. Existing studies are often evaluated on very small cohorts, and performance on small tumors is limited~\cite{cao2023large,korfiatis2023automated,chen2023pancreatic,degand2024validation,alves2025artificial,chen2021radiomic}. The reported sensitivities range from 74.7\% to 85.7\%, but these results are based on only tens of patients. Importantly, these studies are evaluated on diagnostic CT scans, not on prediagnostic scans acquired before clinical detection.

Existing studies have limitations. \textbf{First}, many studies are trained and tested using data from a single population or geographic region~\cite{cao2023large,chen2023pancreatic,degand2024validation,chen2021radiomic}. 
This limits generalizability, as models trained on one population may not perform reliably on patients from different ethnic or clinical backgrounds. \textbf{Second}, reproducibility remains a major challenge. Many studies rely on private datasets for both training and testing, and the trained models are not publicly released~\cite{cao2023large,korfiatis2023automated,qureshi2022predicting,chen2023pancreatic,degand2024validation}. 
As a result, independent validation is difficult, fair comparison across methods is not possible, and reported performance cannot be easily reproduced. \textbf{Third}, many existing approaches focus on patient-level prediction without lesion localization~\cite{qureshi2022predicting,chen2023pancreatic,korfiatis2023automated,chen2021radiomic}. Without localization, it is unclear whether a model bases its decision on pancreatic findings or on unrelated image features. Clinically, the lack of localization also prevents users from knowing where a suspicious lesion is located, which limits interpretability and practical usefulness. 

To address this, we have developed an open-source, automated system, \method\ (\underline{e}arly \underline{P}ancreatic cancer detection with \underline{A}rtificial \underline{I}ntelligence), which detects and localizes small PDACs less than 2 cm in diameter with high accuracy not only in diagnostic scans but also in prediagnostic scans taken 3 to 36 months before clinical diagnosis. This will result in safe and effective detection of early-stage malignancies missed by standard of care diagnostic techniques, and in some cases will enable timely treatment with intent to cure. Our study first evaluated \method\ internally on abdominal contrast-enhanced CT scans, consisting of \internalallpatients\ patients, and then validated \method\ on a larger external multicenter test cohort, consisting of \externalallpatients\ patients, to assess its generalizability to various settings. Furthermore, we study the feasibility of applying \method\ on prediagnostic CT scans, collected 3--36 months before clinical diagnosis, and compare its performance with results from a multi-reader study involving \numofreaders\ radiologists in early and prediagnostic detection of pancreatic cancer.

\section{Results}\label{sec:results}

\begin{figure}
    \centering
    \includegraphics[width=1\linewidth]{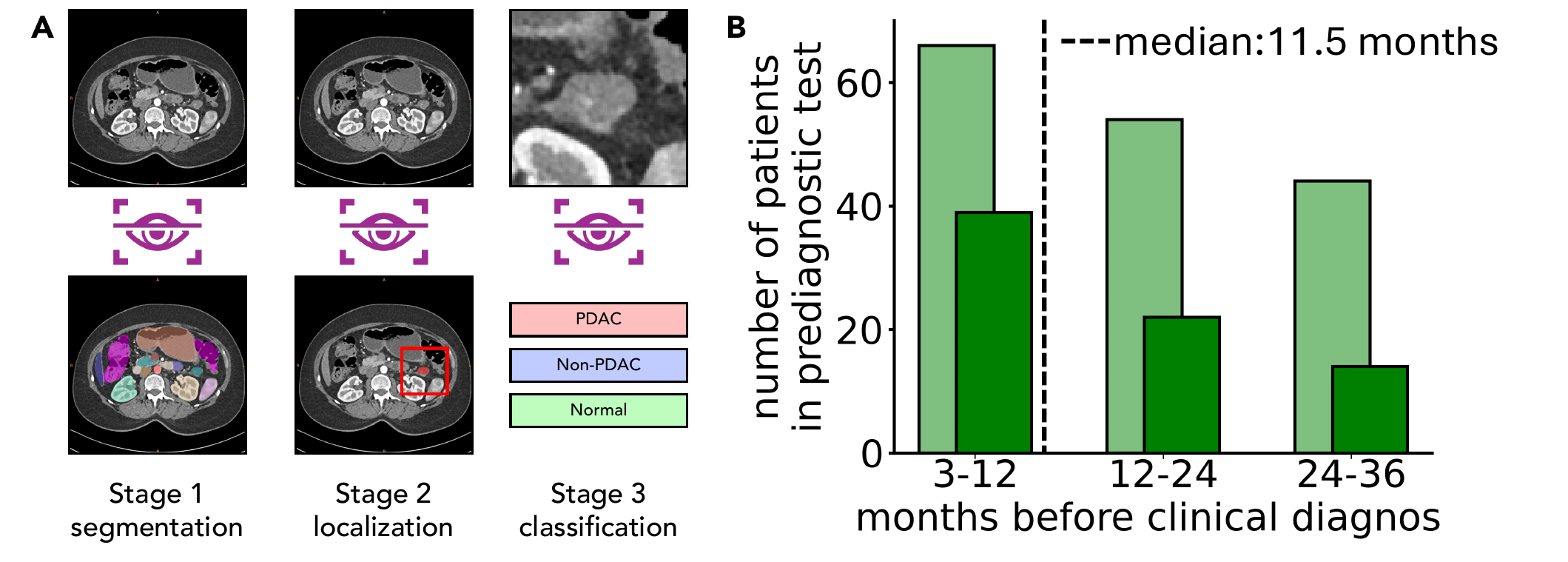}
   \caption{\textbf{Overview and performance of the \method\ system.} 
    \textbf{A.} The \method\ system detects pancreatic ductal adenocarcinoma (PDAC) through a three-stage cascade: Stage 1 segments pancreatic and surrounding anatomy, Stage 2 localizes all potential lesions, and Stage 3 classifies each lesion as PDAC, non-PDAC, or normal. 
    \textbf{B.} Distribution of number of detections in prediagnostic CT scans, showing that \method\ identified PDAC a median of \prediagnostictime\ days before the first clinical diagnosis by radiologists.}

    \label{fig:fig_teaser_ePAI}
\end{figure}

\subsection{The \method\ system}\label{sec:epai_system}

We present \method, an automated system for detecting and localizing PDAC from contrast-enhanced CT scans. \method\ performs three main tasks: detecting the presence of pancreatic lesions, segmenting the lesion, and classifying lesions as PDAC or non-PDAC (\figureautorefname~\ref{fig:fig_teaser_ePAI}A).

\method\ was trained on a dataset of contrast-enhanced abdominal CT scans from 1,598 patients at Johns Hopkins Hospital (JHH) \cite{xia2022felix,park2020annotated}. Patient characteristics are summarized in \tableautorefname~\ref{tab:jhh_dataset}. Ground truth labels were confirmed by surgical pathology for lesion cases and by two-year follow-up for normal controls. \method\ was supervised with pixel-wise annotations of the pancreas and lesions provided by radiologists. To enhance sensitivity, especially for detecting and localizing very small lesions, \method\ was additionally trained on a large set of synthetic lesions with pixel-wise annotations automatically provided by generative AI \cite{li2023early,chen2024towards,lai2024pixel,li2024text}. Further details on the dataset and annotations are provided in the Methods section.

The \method\ system is structured as a three-stage cascade designed to enhance model interpretability, adaptability to varying lesion types and anatomical structures, and performance across tasks of increasing complexity. Stage 1: An nnU-Net model~\cite{isensee2021nnu} was used for anatomical segmentation to localize the pancreas (head, body, tail), surrounding vessels (arteries and veins), and visible dilated ducts. Stage 2: The same nnU-Net model was fine-tuned to detect and localize all potential pancreatic lesions on contrast-enhanced CT scans. It was trained on both real and synthetic lesions to improve sensitivity and included a large cohort of normal scans to reduce false positives and enhance specificity. Stage 3: A classification model analyzed the detected lesions using features like their shape, texture, location, and pancreas morphology (including radiomics features and AI features), allowing for classification among PDAC, non-PDAC lesions and normal.

We mainly evaluated \method\ in detecting and localizing PDACs, with a special focus on small PDACs ($\leq 2$\,cm). The evaluation was conducted on two types of contrast-enhanced CT scans: (1) \textbf{diagnostic scans} where radiologists correctly detected and reported PDACs, and (2) \textbf{prediagnostic scans} where radiologists initially overlooked the PDACs, which were only detected and reported in follow-up scans 3--36 months later. \method\ was an interpretable AI model that directly outputs the segmentation mask of the pancreas, surrounding vessels, dilated ducts and pancreatic lesions if present. Two tasks were evaluated. The first task is PDAC detection: that is, PDAC versus normal, which also includes detection rates stratified by cancer stage. The second task is PDAC localization: a lesion was considered correctly localized only if its position matched the radiologists’ voxel-wise annotations or descriptions in the reports.

\begin{figure}
    \centering
    \includegraphics[width=1\linewidth]{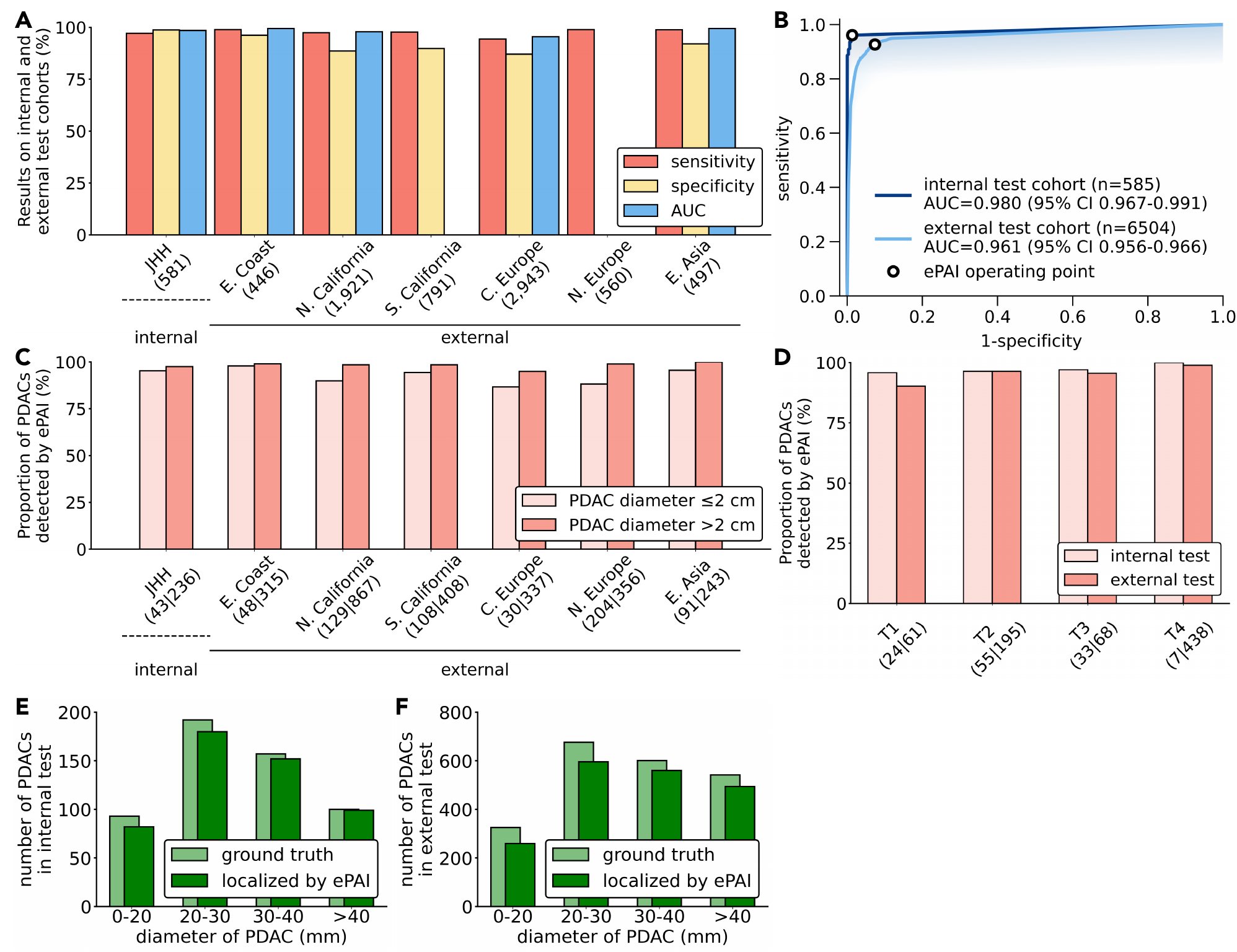}
    \caption{\textbf{Internal and external validation of early PDAC detection with \method.} \textbf{(A)} Sensitivity, specificity, and area under the receiver operating characteristic curve (AUC) of \method\ for detecting all-size PDAC in the internal test cohort and six external regional multicenter test cohorts. \textbf{(B)} Receiver operating characteristic (ROC) curves of \method\ for detecting all-size PDAC across internal and external cohorts, highlighting its generalizability across centers. \textbf{(C)} Sensitivity of \method\ stratified by PDAC lesion size ($\leq 2$\,cm vs. $> 2$\,cm). \textbf{(D)} Sensitivity stratified by PDAC T-stage (T1–T4), showing strong performance even in the earliest stages. Localization performance in internal \textbf{(E)} and external cohorts \textbf{(F)}, evaluated by localization accuracy between the predicted lesion location and radiologists’ voxel-wise annotations or report-based localization, \method\ achieved \internalallsizelocalizationsens\ in internal test cohort while achieved \externalallsizelocalizationsens\ in external test cohort. 
    }
    \label{fig:fig_ePAI_results}
\end{figure}

\subsection{Early detection and localization of PDACs from internal test cohort}\label{sec:small_pdac_detection_internal_evaluation}


Our internal test cohort consisted of \internalallpatients\ patients (279 patients with PDAC, 427 patients with non-PDAC, and \internalnormalpatients\ normal controls) from JHH. These patient labels were confirmed on surgical pathology or a 2-year follow-up. 

We first evaluated PDAC detection at the \textit{patient} level. For all-size PDAC detection, \method\ achieved an area under the receiver operating characteristic curve (AUC) of \internalauc\ (95\% confidence interval (CI): \internalaucrange), a sensitivity of \internalallsizesens\ (95\% CI: \internalallsizesensrange), and a specificity of \internalspec\ (95\% CI: \internalspecrange) as shown in \figureautorefname~\ref{fig:fig_ePAI_results}A--B. The sensitivity for detecting small PDACs (diameter $\leq 2$\,cm) was \internalsmallsens\ (95\% CI: \internalsmallsensrange; n = \internalsmallpatients), precisely detecting PDAC as small as 2 mm; for detecting large PDACs (diameter $> 2$\,cm) was \internallargesens\ (95\% CI: \internallargesensrange; n = \internallargepatients). \method\ achieved a sensitivity of 95.8\% (95\% CI: 79.8--99.3\%; n = 24) for T1 stage, 96.4\% (95\% CI: 87.7--99.0\%; n = 55) for T2 stage, and 97.5\% (95\% CI: 87.1--99.6\%; n = 40) for T3--4 stage (\figureautorefname~\ref{fig:fig_ePAI_results}C). 

We further evaluated PDAC localization at the \textit{lesion} level. Localization was measured whether the model correctly identifies the spatial location of a PDAC lesion. A predicted PDAC was considered correctly localized if it overlapped with the radiologists' voxel-wise annotated tumor mask. We obtained a sensitivity of \internalallsizelocalizationsens\ (95\% CI: \internalallsizelocalizationsensrange; n = 542) for localization of all-size PDACs and \internalsmalllocalizationsens\ (95\% CI: \internalsmalllocalizationsensrange; n = 93) for small PDACs (\figureautorefname~\ref{fig:fig_ePAI_results}E).

\subsection{Early detection and localization of PDACs from external multicenter test cohorts}\label{sec:small_pdac_detection_external_evaluation}

To assess the generalizability of \method\ to different patient populations and imaging protocols, we validated our model on external multicenter (n = \externalcenters) test cohorts, which consisted of contrast-enhanced abdominal CT scans of \externalallpatients\ patients (\externalsmallpatients\ with small PDAC, \externallargepatients\ with large PDAC, and \externalnormalpatients\ normal controls) from North America, Europe, and Asia. The patient labels were confirmed by histopathology reports or a 2-year follow-up visit diagnosis. 

\method\ achieved an AUC of \externalauc\ (95\% CI: \externalaucrange), sensitivity of \externalallsizesens\ (95\% CI: 96.1--97.6\%), and specificity of \externalspec\ (95\% CI: \externalspecrange) for all size PDAC detection. For the small PDAC patient subgroup (diameter $\leq 2$\,cm), the sensitivity was \externalsmallsens\ (95\% CI: \externalsmallsensrange; n = \externalsmallpatients). For the large PDAC patient sub-group (diameter $> 2$\,cm), the sensitivity was \externallargesens\ (95\% CI: \externallargesensrange; n = \externallargepatients). The PDAC detection results for each center are shown in \figureautorefname~\ref{fig:fig_ePAI_results}A--C. \method\ achieved a sensitivity of 90.2\% (95\% CI: 80.2--95.4\%; n = 61) for T1, 96.4\% (95\% CI: 92.8--98.3\%; n = 195) for T2, and 98.4\% (95\% CI: 96.9--99.2\%; n = 506) for T3--4. We obtained a sensitivity of \externalallsizelocalizationsens\ (95\% CI: \externalallsizelocalizationsensrange; n = 2,404) for localization of all-size PDACs and \externalsmalllocalizationsens\ (95\% CI: \externalsmalllocalizationsensrange; n = 325) for small PDACs (\figureautorefname~\ref{fig:fig_ePAI_results}F).


\begin{figure}
    \centering
    \includegraphics[width=1\linewidth]{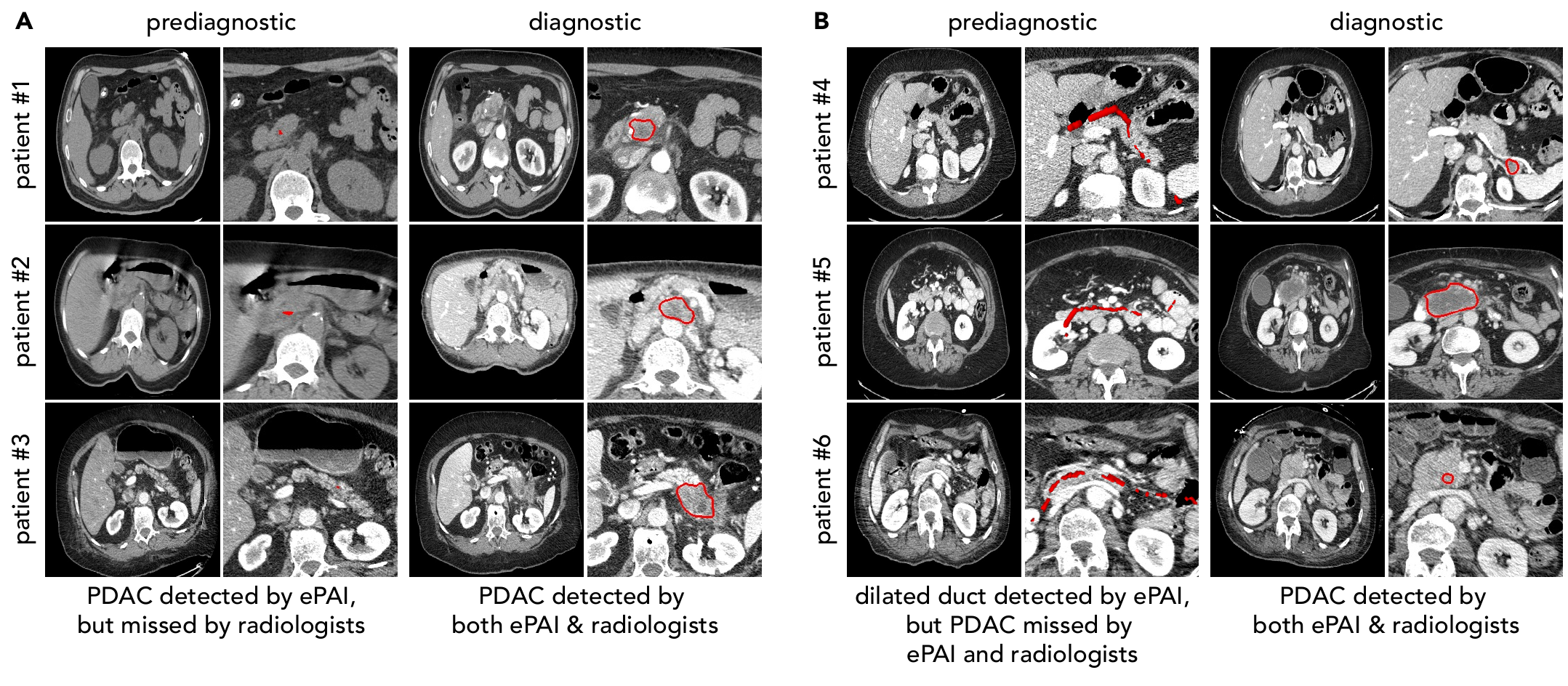}
    \caption{\textbf{Earlier PDAC detection in prediagnostic CT scans.} Representative paired prediagnostic (left) and diagnostic (right) contrast-enhanced CT images from six patients in the external multicenter prediagnostic cohorts (3–36 months before first clinical diagnosis). Red contours indicate \method\ predictions or radiologist-confirmed tumor regions on the diagnostic scans. \textbf{(A)} Patients \#1–\#3 show \emph{direct prediagnostic detection}: \method\ marks small, subtle abnormalities on the prediagnostic scans (small red contours), and the diagnostic scans later show an obvious PDAC at the same location (red contours), detected by both \method\ and radiologists. \textbf{(B)} Patients \#4–\#6 show \emph{indirect prediagnostic cues}: \method\ highlights secondary changes on the prediagnostic scans, most notably pancreatic duct dilatation or cutoff (red 3D rendering), even when no clear mass is visible to \method\ or radiologists. The paired diagnostic scans later show a clear PDAC (red contour) detected by both \method\ and radiologists. Together, these examples illustrate two common prediagnostic patterns---early subtle lesions and secondary ductal changes before an obvious mass---and potentially support the use of \method\ for earlier PDAC review.}
    \label{fig:prediagnostic_ct}
\end{figure}

\subsection{Prediagnostic detection and localization of PDACs from external multicenter test cohorts}\label{sec:prediagnostic_pdac_detection_external_evaluation}

To investigate whether \method\ can detect PDAC before radiologists, we tested it on CT scans acquired 3--36 months prior to the first clinical diagnosis by radiologists. These scans are called prediagnostic scans. External multicenter (n = 3) test cohorts were collected from North America and Eastern Asia, including \prediagnosticallpatients\ patients with both prediagnostic and diagnostic CT scans. This cohort was eventually diagnosed with PDAC. Unfortunately, in practice, radiologists 
failed to detect PDACs in 100\% of these prediagnostic scans, indicating that theses lesions were retrospectively visible but initially overlooked. 
The diagnostic scans from this same cohort showed advanced disease, with an average stage of 2.67 and a mean tumor diameter of $3.1 \pm 1.5$ cm. 
The normal controls remained the same as those used for external validation of early PDAC detection (Section \ref{sec:small_pdac_detection_external_evaluation}).

Without additional training or parameter adjustment on prediagnostic CT scans, \method\ successfully detected PDAC in \prediagnosticdetected\ out of \prediagnosticallpatients\ patients, with a median leading time of \prediagnostictime\ days before first clinical diagnosis by radiologists. Localization accuracy was assessed by comparing predicted PDAC sites in prediagnostic scans with the confirmed PDAC locations in diagnostic scans (\figureautorefname~\ref{fig:prediagnostic_ct}A). Localization was considered correct when the predicted region corresponded to the same pancreatic segment (head, body, or tail) as in the diagnostic scan. \method\ correctly localized PDAC in \prediagnosticlocalized\ of the patients whose tumors were directly detected.
Moreover, \method\ identified secondary morphological changes like pancreatic-duct dilation in \prediagnosticductdetected\ of the patients whose tumors were not directly detected (Figure~\ref{fig:prediagnostic_ct}B). These findings indicate that \method\ can recognize early structural alterations preceding overt tumor visibility.

\begin{figure}
    \centering
    \includegraphics[width=1\linewidth]{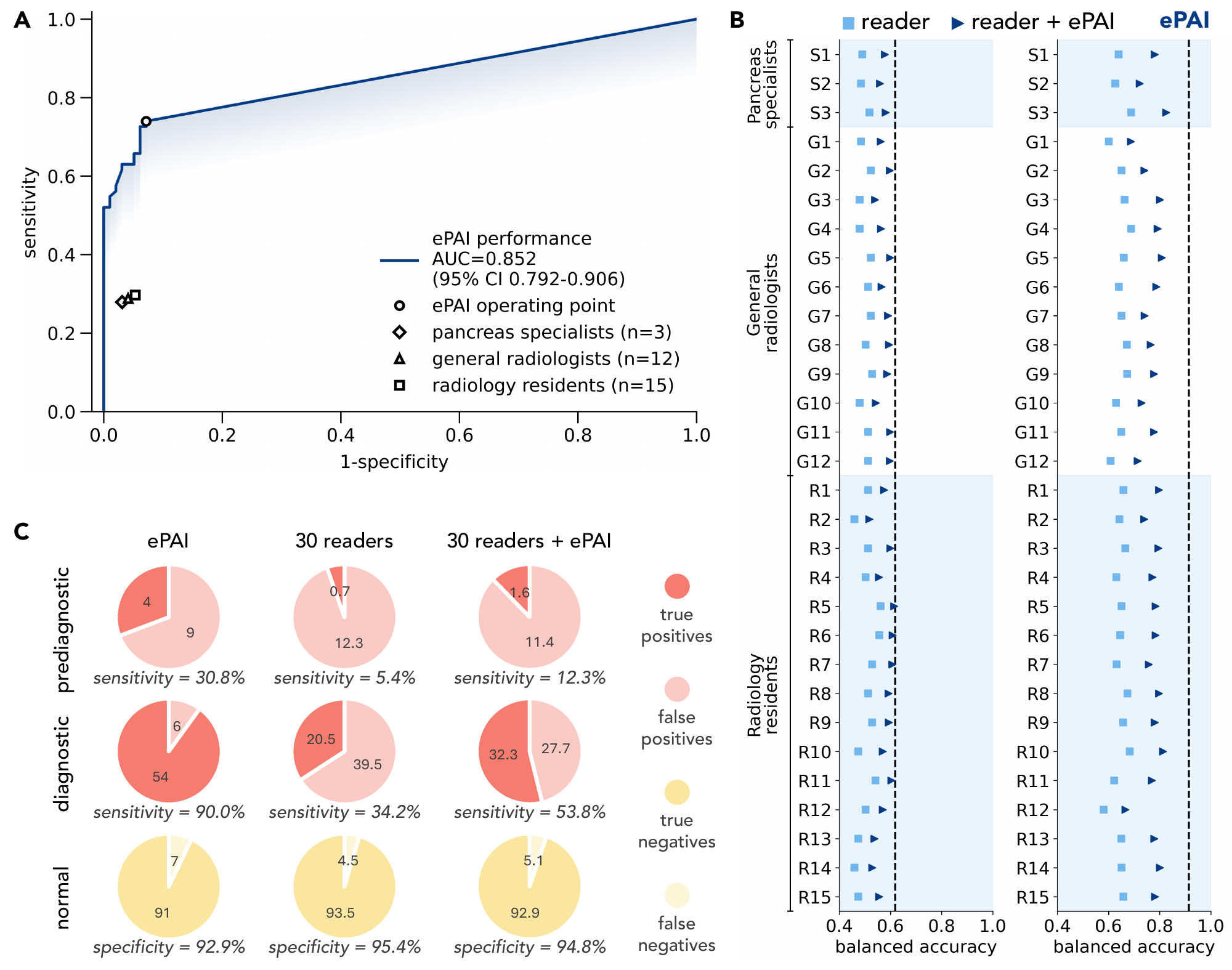}
    \caption{\textbf{Multi-reader study.} (A) Comparison between \method\ and 30 readers with different expertise, evaluated on prediagnostic, diagnostic, and normal CT scans. \method\ shows substantially higher sensitivity in both prediagnostic and diagnostic settings while maintaining high specificity on normal controls. (B) Balanced accuracy of individual readers (pancreatic imaging specialists, general radiologists, and radiology residents) compared with \method\ and themselves with assistance of \method. (C) Sensitivity, specificity of \method, readers, and reader-plus-\method\ combinations for prediagnostic, diagnostic, and normal cohorts.}
    \label{fig:reader_study}
\end{figure}

\subsection{Multi-reader study and comparison between ePAI and radiologists}\label{sec:reader_studies}

We conducted a multi-reader study to compare performance of \method\ with radiologists in detecting early PDAC less than 2 cm. A total of \numofreaders\ readers (median years of experience: 4, mean: 6.1, s.d.: 7.5, range: 2--37) participated in the study. The readers included pancreas specialists (n = 3), general radiologists (n = 12), and radiology residents (n = 15) recruited from Johns Hopkins Medicine (JHMI), City of Hope (CoH), the University of California, San Francisco (UCSF), the University of Illinois Urbana–Champaign (UIUC), Shandong Provincial Qianfoshan Hospital (SFMU), and Qilu Hospital of Shandong University (QLH). 

Both \method\ and human readers reviewed a mixed cohort including 13 prediagnostic and 63 diagnostic CT scans with early PDACs, as well as 99 normal CT scans with at least 2 years follow-up as controls. This proportion is unknown to the readers beforehand. Readers reviewed all the CT scans and rated each case as PDAC or normal. The results in \figureautorefname~\ref{fig:reader_study} showed that \method\ outperformed 30 radiologists by \outperformreadersens\ (95\% CI: \outperformreadersensrange) in sensitivity while maintaining more than 92.0\% specificity in detecting PDACs early and prediagnostic. Specifically, for diagnostic scans, \method\ achieved a sensitivity of 90.0\%, markedly higher than the average of 34.2\% among the 30 readers. For prediagnostic scans, \method\ detected 30.8\% of PDACs that were initially missed by radiologists, compared with only 5.4\% detected by readers after they re-reviewed these scans. When assisted by \method, radiologists improved sensitivity by \deltareaderplusepaidiagnocticsens\ and \deltareaderplusepaiprediagnocticsens\ in diagnostic and prediagnostic scans, respectively. 
Collectively, these results demonstrate that \method\ substantially outperforms human readers in early and prediagnostic PDAC detection while achieving a similar specificity on normal cases.


\section{Discussion}
\label{sec:discussion}

We observed strong performance of \method\ for PDAC detection and localization, with robust generalization across external cohorts and substantial improvement over radiologists. With 94.4\%--98.5\% sensitivity and 96.6\%--99.5\% specificity, \method\ reached clinically meaningful positive predictive value (PPV) in enriched high-risk populations with PDAC prevalence of 3.6\%--4.4\% \cite{sharma2018model,dbouk2022multicenter,blackford2024pancreatic}, yielding a PPV of 50.9\%--90.0\%. In this setting, screening 100,000 individuals would identify approximately 3,398--4,334 true positives among 4,812--6,676 AI-flagged cases. In contrast, PDAC is rare in the general population. At low-risk prevalence levels (0.014\%--0.03\%) \cite{bray2024global,sung2021global,siegel2025cancer}, screening 100,000 individuals would yield only 13--30 true positives among 530--3,413 positive predictions (PPV: 0.4\%--5.6\%), meaning that the vast of majority of flagged cases would be false positives.

Recent AI studies on pancreatic cancer detection have only been tested on single population or very limited geographic regions \cite{cao2023large,korfiatis2023automated,chen2023pancreatic,degand2024validation,alves2025artificial,chen2021radiomic}, restricting evaluation of generalizability. In contrast, we conducted large-scale external validation across six independent regional cohorts spanning North America, Europe, and Asia, including a national registry from Poland aggregating data from more than 60 hospitals. This design enables robust assessment across heterogeneous patient populations, scanner vendors, and acquisition protocols. Detection of small tumors $\leq 2$\,cm is clinically important because these tumors are more likely to be surgically resectable. Unlike most prior work, which either did not report performance on small tumors \cite{chen2021radiomic} or included only a few such cases \cite{cao2023large}, we conducted extensive subgroup analyses on 653 patients with small tumors, demonstrating that \method\ maintains high sensitivity for early-stage PDACs. The observed robustness is supported by the training design and supervision signals used in this study, including synthetic tumor augmentation and rich voxel-wise annotations that provide anatomical context beyond the tumor itself (e.g., pancreas and pancreatic sub-regions, pancreatic duct, adjacent vessels, and surrounding organs).

Compared with PDAC detection, tumor localization has received far less systematic evaluation in the literature. Most prior studies focus on patient-level classification and, at most, provide qualitative visualization tools such as saliency maps or heatmaps, often assessed on a limited number of cases without explicit anatomical ground truth. 
In this study, we define localization as correctly identifying the pancreatic region (head, body, or tail) containing the tumor, and we evaluate this capability using voxel-wise tumor annotations in the internal cohort and anatomically defined pancreatic segments in external cohorts. \method\ correctly localizes more than 90\% of detected tumors to the appropriate pancreatic region. This represents one of the most comprehensive quantitative evaluations of PDAC localization to date. Localization is clinically important because it reduces radiologist search burden, mitigates perceptual errors, and enables actionable downstream decisions, such as targeted endoscopic ultrasound–guided biopsy. More broadly, localization distinguishes imaging-based approaches from non-imaging early-detection methods, such as blood-based or molecular assays, which may indicate elevated cancer risk but cannot specify the organ or anatomical site of disease. By providing spatially resolved information, \method\ bridges the gap between early cancer suspicion and organ-specific clinical workup.

A central contribution of this study is the evaluation of \method\ on prediagnostic CT scans acquired between 3 and 36 months before clinical diagnosis. This is one of the most clinically important---and technically difficult---settings for early PDAC detection: abnormalities are often subtle, indirect, or not yet a discrete mass. In our external prediagnostic cohort (159 patients with paired prediagnostic and diagnostic scans), radiologists did not report PDAC in any prediagnostic scan at the time of care, whereas \method\ detected PDAC in 75/159 patients with a median lead time of 347 days and localized the predicted site to the correct pancreatic segment (head/body/tail) in 75\% of detected cases. These findings highlight two distinct clinical scenarios that explain why PDAC is missed in practice. First, the radiologist may be “correct” under a lesion-centric standard: the scan may not contain a clearly actionable tumor, but may show early or non-specific precancerous or high-risk morphologic changes (e.g., subtle focal abnormalities) that would justify risk stratification and closer surveillance rather than immediate biopsy or intervention. Second, the radiologist may be “incorrect” in the sense of perceptual or cognitive miss: an already visible abnormality is overlooked or dismissed, leading to delayed workup and subsequent imaging or surgery once the disease becomes obvious. Additional contributors include time pressure in routine abdominal CT interpretation, attention competition from non-pancreatic findings, protocol and image-quality variability (contrast timing, slice thickness, partial pancreatic coverage), lack of explicit prompts to scrutinize the pancreas when the indication is unrelated, and the fact that early PDAC may present primarily through secondary signs rather than a conspicuous mass. By flagging suspicious regions and related structural cues (including pancreatic-duct changes), \method\ can shift care from passive detection to proactive risk-aware follow-up—either prompting earlier targeted review when a lesion is already present, or supporting earlier surveillance when only subtle high-risk changes exist.

The multi-reader study underscores the persistent difficulty of detecting early PDAC. Radiologists across training levels showed low sensitivity for small and prediagnostic tumors, consistent with performance reported in the literature. 
AI assistance improved sensitivity for all readers, including those without subspecialty experience, and maintained high specificity on normal scans. These findings align with established benefits of AI-augmented interpretation in other screening domains, such as lung nodules and breast cancer. They support the emerging concept that opportunistic AI review of abdominal CT scans may enable earlier PDAC detection by drawing attention to subtle abnormalities that would otherwise be missed.

These results carry several implications for opportunistic PDAC detection. First, the consistent performance of \method\ across multiple regional centers---including, but not limited to, Northern California---suggests feasibility for broad deployment. 
Second, strong localization performance may help radiologists confirm or refute abnormalities quickly without extended search. Third, the ability to detect PDAC prior to clinical diagnosis indicates potential for earlier identification, though further refinement and prospective validation are required. Given the low prevalence of PDAC, integrating \method\ with clinical risk models (e.g., ENDPAC
) or laboratory markers to identify higher-risk populations may reduce false positives while maximizing benefit. Evaluating downstream consequences of AI-generated alerts, including follow-up imaging and invasive procedures, will be essential.

In one of our six multicenter external test cohorts, Northern California dataset, we made it scale by using large language models (LLMs) to assist in identifying normal controls, path-proven PDAC cases, and prediagnostic scans, followed by expert verification. This workflow highlights the value of LLMs for accelerating data curation, a major bottleneck in developing medical imaging AI systems, and aligns with emerging evidence that agentic LLM pipelines can support large-scale clinical dataset assembly.

Our study has limitations. Clinical information was incomplete for some external sites, limiting certain subgroup analyses. The multi-reader study did not measure interpretation time, preventing direct assessment of workflow efficiency. Finally, performance varied across centers, likely reflecting heterogeneity in scanners, protocols, and contrast timing. Future work should address these domain shifts and evaluate \method\ in prospective clinical settings. 

\section{Methods}
\subsection{Dataset description}\label{sec:dataset_description}
This multicenter retrospective study comprised five patient cohorts: \textbf{(I)} an internal training cohort of diagnostic CT scans for model development; \textbf{(II)} an internal test cohort of diagnostic CT scans for performance evaluation; \textbf{(III)} an external multicenter diagnostic CT cohort (n = \externalcenters) for generalization assessment; \textbf{(IV) }an external multicenter prediagnostic CT cohort (n = \prediagnosticcenters); and \textbf{(V)} an external multicenter reader-study cohort including diagnostic, prediagnostic, and normal scans. Across all cohorts, PDAC was confirmed by pathology, including histopathology of surgical specimens or cytology when surgery was not performed, followed the 2019 WHO Classification of Tumors~\cite{nagtegaal20192019}. Patients with mixed neoplasms (e.g., PDAC with neuroendocrine or cystic components) were excluded. Normal controls were confirmed to be free of pancreatic disease, and patients with acute pancreatitis or prior abdominal treatment were excluded.

\smallskip\noindent\textbf{Internal training and testing cohorts (Cohorts I \& II)} comprised 2,519 patients, each patient with dual-phase contrast-enhanced CT scans acquired in the arterial and portal-venous phases, resulting in a total of 5,085 annotated CT scans. All scans were pancreatic-protocol CTs acquired on Siemens multidetector CT scanners. This retrospective study was approved by the Johns Hopkins Hospital Institutional Review Board under IRB00403268. Diagnostic CT scans included 3,440 scans acquired between 2003 and 2020 and identified from clinical, radiology, and pathology databases. Among these, 864 patients had pathology-confirmed PDAC, including 576 patients in Cohort I and 288 patients in Cohort II, with PDAC tumors stratified by size ($\leq 2$\,cm vs.\ $> 2$\,cm) as summarized in Table~\ref{tab:jhh_dataset}. The remaining diagnostic scans corresponded to non-PDAC pancreatic diseases. Normal controls consisted of 1,645 CT scans from 836 renal donor patients without pancreatic tumors, including 533 patients in Cohort I and 303 patients in Cohort II. To minimize the likelihood of undiagnosed pancreatic disease, 99\% of renal donor scans were acquired before 2010. Both arterial and portal-venous phases were independently used for model training.

\smallskip\noindent\textbf{Lesion and pancreas annotation.}  
The entire three-dimensional pancreas and PDAC tumors were manually segmented by five trained annotators using commercial segmentation software 3D Slicer for Cohorts I and II. For subjects with dual-phase CT scans, pancreas and PDAC tumors were independently annotated in both arterial and portal-venous phases by a single annotator with around 2 years in experience post residency. All annotations were subsequently reviewed and verified by one of three board-certified radiologists who were not involved in the initial segmentation. To mitigate the impact of annotation errors and inter- or intra-observer variability on model training and evaluation, a dedicated quality control and error-correction procedure was applied. Potential annotation errors were first identified through expert visual inspection and then systematically screened using in-house software to detect major inconsistencies, such as missing slices or incomplete organ coverage within the region of interest. Radiologist re-review was conducted to correct errors in the ground truth, including missed small PDAC tumors and slight inaccuracies in their locations.

\smallskip\noindent\textbf{External multicenter test cohorts (Cohorts III).}  
Independent evaluation of model generalization was performed using external multicenter test cohorts (Cohort III) comprising \externalallpatients\ patients from the North America, Europe, and Eastern Asia, including \externalpdacpatients\ PDAC cases and \externalnormalpatients\ normal controls (Tables~\ref{tab:external_diagnostic}-\ref{tab:external_normal}). Data sources were grouped by geographic region. All external cohorts were fully held out from model training and validation. The East Coast (North America) cohort included 446 patients, comprising 363 diagnostic PDAC cases and 80 normal controls. Diagnostic cases were confirmed by histopathology. Normal controls were defined by the absence of pancreatic malignancy or major abdominal pathology. Normal scans were acquired exclusively in the portal-venous phase, with a median in-plane spacing of 0.86\,mm and a slice thickness of 1.0\,mm. The mean age of normal subjects was $46.8 \pm 16.7$ years, with 33.8\% female and 66.2\% male participants. The Northern California (North America) comprised 1,921 patients, including 996 diagnostic PDAC cases and 952 normal controls. Diagnostic cases had a mean age of $65.4 \pm 10.7$ years and included 129 PDAC tumors $\leq 2$\,cm (13.0\%) and 867 PDAC tumors $> 2$\,cm (87.0\%). Normal controls had a mean age of $63.5 \pm 14.8$ years, with 46.2\% female and 42.5\% male subjects. Most normal scans were acquired in the portal-venous phase (90.2\%), with a median slice thickness of 1.25\,mm and a median in-plane spacing of 0.74\,mm. The Southern California (North America) cohort included 791 patients, consisting of 521 diagnostic PDAC cases and 526 normal controls. Diagnostic cases included 110 PDAC tumors $\leq 2$\,cm (21.1\%) and 411 PDAC tumors $> 2$\,cm (78.9\%), with a mean age of $66.9 \pm 10.6$ years. Normal controls had a mean age of $59.4 \pm 18.1$ years, with 48.3\% female and 51.7\% male subjects. CT scans exhibited increased heterogeneity in acquisition, with a median slice thickness of 2.55\,mm and mixed contrast phases, including portal-venous (62.2\%) and arterial (24.9\%). The Central Europe cohort comprised 2,994 patients, including 367 diagnostic PDAC cases and 2,627 normal controls. The normal cohort comprises 1,227 females (46.7\%), and 1,400 males (53.3\%), with a mean age of $59.2 \pm 15.6$ years. All diagnostic PDAC tumors were pathology-confirmed, including 30 PDAC tumors $\leq 2$\,cm (8.2\%) and 337 PDAC tumors $> 2$\,cm (91.8\%). The PDAC cohort comprises 177 females (48.2\%), and 190 males (51.8\%), with a mean age of $58.9 \pm 13.9$ years. For normal controls, scanner vendors included Toshiba (4.1\%), Siemens (18.9\%), Philips (18.9\%), GE (52.3\%), and Canon (5.8\%), while for PDAC cases Toshiba (2.8\%), Siemens (33.2\%), Philips (16.5\%), GE (46.0\%), and Canon (1.5\%). Both normal controls and diagnostic PDAC cases were exclusively of European ancestry. CT scans of normal control cohort were predominantly acquired in the portal-venous phase, with a median slice thickness of 1.25\,mm and a median in-plane spacing of 0.76\,mm, while diagnostic PDAC cohort were predominantly acquired in both arterial (48.5\%) and portal-venous (51.5\%) phases, with a median slice thickness of 1.25\,mm and a median in-plane spacing of 0.77\,mm. The Northern Europe cohort consisted of 560 diagnostic PDAC patients only and did not include true normal controls. PDAC tumor size distribution included 204 PDAC tumors $\leq 2$\,cm (36.4\%) and 356 PDAC tumors $> 2$\,cm (63.6\%). All cases were histopathology-confirmed. CT scans were acquired exclusively in the portal-venous phase, with a median slice thickness of 2.0\,mm and a median in-plane spacing of 0.73\,mm. Scanner vendors included Philips (41.4\%), Siemens (29.4\%), Toshiba (23.5\%), and GE (4.3\%). The Eastern Asia cohort included 217 patients, comprising 112 diagnostic PDAC cases and 105 normal controls. Diagnostic cases had a mean age of $60.9 \pm 9.4$ years and included 91 PDAC tumors $\leq 2$\,cm (27.2\%) and 243 PDAC tumors $> 2$\,cm (72.8\%). Normal controls had a mean age of $60.6 \pm 11.1$ years. Diagnostic scans included arterial (33.5\%) and portal-venous (33.2\%) phases, whereas normal scans were predominantly portal-venous (60.1\%). The median slice thickness for diagnostic scans was 1.0\,mm.

\noindent\textbf{Prediagnostic and reader study cohort (Cohorts IV and V).} Using electronic medical records (EHR) together with LLMs, we identified patients with biopsy-confirmed PDAC diagnosed between January 2006 and December 2020 in Northern California, United States ($n=522$). We then retrieved abdominal CT examinations acquired for non-pancreatic indications 3–36 months before the index PDAC diagnosis, yielding 48 prediagnostic patients. We applied the same pipeline at two external hospitals in Eastern Asia, identifying an additional 111 prediagnostic patients. Prediagnostic examinations were required to have been interpreted as negative for PDAC in routine care, that is, no reported suspicion of pancreatic tumor. For patients with multiple eligible examinations, we included all scans acquired within 36 months before diagnosis that met the prediagnostic criteria. Each CT was re-reviewed by one of two radiologists (2–4 years post-residency experience) to confirm optimal image quality (e.g., minimal motion artifacts), contrast enhancement, and no evidence of pancreatitis, focal lesions, or common bile duct stents.

\subsection{AI model: \method}\label{sec:ai_model_epai}

\method\ is an automated system for detecting and localizing PDAC from contrast-enhanced abdominal CT scans, including prediagnostic scans acquired 3--36 months before the first clinical diagnosis by radiologists. To improve interpretability and support lesion-level localization, \method\ is designed as a three-stage cascade that outputs spatial predictions rather than only patient-level scores.

\noindent\textbf{Three-stage cascade.} Stage~1 performs anatomical segmentation to localize the pancreas (head, body, tail), surrounding vessels (arteries and veins), and visible dilated ducts. Stage~2 detects and localizes all potential pancreatic lesions. Stage~3 classifies each detected lesion as PDAC, non-PDAC, or normal. This separation of segmentation, localization, and classification improves transparency and enables direct inspection of the predicted lesion location.

\noindent\textbf{Training supervision and label confirmation.} \method\ was trained on 1,598 contrast-enhanced 
abdominal CT scans from JHH. Ground-truth 
labels were confirmed by pathology 
for lesion cases and by two-year follow-up for normal controls. Radiologists provided pixel-wise annotations of the pancreas and lesion to supervise Stage~1 and Stage~2.

\noindent\textbf{Synthetic lesions for small-tumor sensitivity.} To increase sensitivity for small PDACs, \method\ was additionally trained on a large set of synthetic lesions with pixel-wise annotations automatically provided by generative AI. Synthetic lesions were used to enrich rare appearances and to expose the localization model to tumors near the lower limit of visibility. 


\noindent\textbf{Evaluation cohorts and clinical comparison.} We evaluated \method\ on two clinically defined settings: (i) diagnostic scans, in which radiologists detected and reported PDAC, and (ii) prediagnostic scans, in which PDAC was initially overlooked and only detected in follow-up scans 3--36 months later. We assessed PDAC detection (PDAC versus normal) at the patient level and PDAC localization at the lesion level. Localization was considered correct when the predicted lesion location matched radiologists' pixel-wise annotations or report-based localization. We further compared \method\ with radiologists, and measured the effect of assistance by \method, using a multi-reader study.

\subsection{Evaluation metrics}\label{sec:evaluation_metrics}

\noindent\textbf{PDAC detection.}
We evaluate PDAC detection as a binary classification task where ePAI determines whether each CT scan has a PDAC or not. For interpretability, we perform detection through semantic segmentation. By segmenting the tumor, ePAI provides localization, allowing radiologists to better understand and verify AI decisions. We consider that ePAI detected PDAC in a CT scan if the largest PDAC region segmented inside the pancreas has a maximum per-voxel PDAC probability greater than 50\%. This definition of detection was used to compute the area under the receiver operating characteristic curve (AUC), sensitivity, specificity, accuracy, and balanced accuracy. We also stratified performance by T stage (T1--T4). Importantly, ePAI differentiates between PDAC and non-PDAC lesions, both present in our test sets.

\noindent\textbf{PDAC localization.}
We considered a lesion correctly localized if the AI-generated tumor segmentation mask overlapped with the radiologist-drawn tumor segmentation mask. For test datasets without radiologist-drawn tumor masks, we considered a lesion correctly localized if the AI segmented it within the pancreatic sub-segment (head, body, or tail) that matched the lesion location described in the radiology report. We then computed the accuracy as the number of correctly localized lesions divided by the total number of lesions in the test dataset.


\noindent\textbf{Ablation studies.}
We performed two ablation studies to quantify the contribution of key design choices in \method. First, for lesion localization (Stage~2), we compared \method\ with a standalone nnU-Net trained directly for pancreatic lesion 
segmentation without Stage~1 initialization; lesion presence was defined by whether the model predicted any non-zero lesion voxels. Second, for lesion classification (Stage~3), we compared the full lesion-conditioned classifier with an image-level baseline that classifies the original CT directly (without using Stage~2 candidate lesions or lesion-level features). 


\noindent\textbf{Reader studies.} We conducted a two-session reader study to assess radiologists’ performance for pancreatic lesion detection 
and PDAC diagnosis on prediagnostic CT, and to test whether \method\ improves reader performance when used as an assistive tool. The two sessions were separated by a washout period of at least six month. A total of \numofreaders\ readers from 5 institutions participated, including 3 pancreatic imaging specialists, 12 general radiologists and 15 radiology residents. Readers had a mean of 6.1 years of experience (range, 2–37 years) and reported reviewing a mean of 575 pancreatic CT examinations in the year before the study. Each reader interpreted 175 UCSF CT examinations, including 13 prediagnostic and 63 diagnostic CT scans with early PDAC, and 99 normal controls; readers were not informed of the class distribution. Readers were provided with the CT scan and patient age and sex. They answered three questions for each case: (1) whether a pancreatic lesion was present (yes/no); (2) if present, the lesion location within the pancreas; and (3) if present, whether the lesion was suspicious for PDAC, suspicious for a non-PDAC lesion or indeterminate.

In Session~1, readers reviewed cases using videos for visualization, without time constraints, and recorded their responses. In Session~2, readers interpreted the same cases with access to \method’s outputs (e.g., PDAC segmentation masks). We quantified changes in reader performance between sessions. All PDAC and normal scans included in the reader study had optimal image quality 
and no signs of pancreatitis. Scans with biliary or pancreatic duct stents, particularly common bile duct stents, were excluded based on visual assessment by two radiologists.

\noindent\textbf{Interpretability of the AI model.} \method\ is interpretable at both the patient and lesion levels. For each examination, it outputs (i) a patient-level probability of abnormality, (ii) a lesion subtype prediction when abnormal, and (iii) voxel-wise segmentation masks of the detected lesion. In addition, Stage~1 provides multi-organ and vessel segmentations that localize the pancreas and surrounding anatomy, including visible ductal structures. These anatomical priors provide spatial context for \method's outputs and facilitate inspection of clinically relevant imaging cues. We quantified the spatial correspondence between predicted and reference segmentations using the localization sensitivity and the 95th percentile Hausdorff distance (HD95). To assess whether \method\ highlights clinically meaningful cues beyond the lesion itself, we also examined its predictions in the context of associated imaging findings and risk-related features, including pancreatic duct dilatation and related peripancreatic anatomical changes captured by the multi-organ and vessel segmentations.

\noindent\textbf{Statistical analysis.}
For two group comparisons, we report the area under the receiver operating characteristic curve (AUC), sensitivity, specificity, positive predictive value (PPV), accuracy and F1 score. AUC was computed from model logits using \texttt{sklearn.metrics.roc\_auc\_score}. The remaining metrics were computed from thresholded predictions using \texttt{sklearn.metrics} (for example, \texttt{confusion\_matrix} and \texttt{precision\_re\-call\_fscore\_support}). We derived 95\% confidence intervals (CIs) for sensitivity, specificity, accuracy and PPV using Wilson score intervals implemented with \texttt{statsmodels.stats.proportion.proportion\_confint} (\texttt{method='wilson'}). We derived 95\% CIs for AUC and F1 score using a nonparametric bootstrap with 1{,}000 resamples. All analyses were performed in Python using NumPy, SciPy, scikit-learn and statsmodels. 

\bmhead{Acknowledgments}
This work was supported by the National Institutes of Health (NIH) under Award Number R01EB037669, the Lustgarten Foundation for Pancreatic Cancer Research, and the Center for Biomolecular Nanotechnologies, Istituto Italiano di Tecnologia (73010, Arnesano, LE, Italy). We would like to thank the Johns Hopkins Research IT team in \href{https://researchit.jhu.edu/}{IT@JH} for their support and infrastructure resources where some of these analyses were conducted, especially \href{https://researchit.jhu.edu/research-hpc/}{DISCOVERY HPC}; thank the HPC infrastructure and the Support Team at Fondazione Istituto Italiano di Tecnologia. 

\bibliography{refs,zzhou}

\clearpage

\begin{appendices}
\section{Supplementary Tables}

\begin{table*}[h]
\centering
\caption{Characteristics of external \emph{diagnostic} CT datasets across centers.}
\resizebox{\textwidth}{!}{
\begin{tabular}{lcccccc}
\toprule
\textbf{Variable}  & \textbf{E.Coast} & \textbf{N.California} & \textbf{S.California} & \textbf{C.Europe} & \textbf{N.Europe} & \textbf{E.Asia} \\
\midrule
CT scan & 363 & 996 & 521 & 712 & 560 & 334 \\
Patient & 363 & 996 & 521 & 367 & 560 & 112 \\
Age, mean (SD) & -- & 65.4 (10.7) & 66.9 (10.6) & 58.9 (13.9) & 68.0 (10.0) & 60.9 (9.4)\\
\quad Unknown (no.) & -- & 762 & -- & -- & -- & -- \\
\midrule
\textbf{Sex} & & & & & & \\
\quad Female, no. (\%) & -- & 486 (48.8) & 260 (49.3) & 177 (48.2) & 270 (48.2) & 35 (31.3) \\
\quad Male, no. (\%)  & -- & 509 (51.1) & 267 (50.7) & 190 (51.8) & 290 (51.8) & 77 (68.8) \\
\quad Unknown (no.)  & -- & 1 (0.1) & -- & -- & -- & -- \\
\midrule
In-plane spacing, mm (IQR) & 0.79 (0.70, 0.92) & 0.74 (0.70, 0.81) & 0.99 (0.74, 1.25) & 0.77 (0.71,0.83)  & 0.73 (0.68, 0.78) & 0.74 (0.70, 0.79) \\
Slice thickness, mm (IQR)  & 2.50 (1.00, 5.00) & 1.25 (1.25, 1.25) & 1.32 (0.51, 2.00) & 1.25 (0.80, 1.50) & 2.00 (1.00, 3.00) & 1.00 (1.00, 1.25) \\
\midrule
\textbf{Scanner, no. (\%)} & & & & & & \\
\quad TOSHIBA  & -- & -- & -- & 20 (2.8) & 131 (23.5) & -- \\
\quad SIEMENS  & -- & -- & -- & 236 (33.2) & 165 (29.4) & -- \\
\quad Philips  & -- & -- & -- & 117 (16.5) & 232 (41.4) & -- \\
\quad GE  & -- & -- &  & 328 (46.0) & 24 (4.3) & -- \\
\quad Canon  & -- & -- & -- & 11 (1.5) & 6 (1.1) & -- \\
\quad Unknown  & -- & -- & -- & -- & 2 (0.4) & -- \\
\midrule
\textbf{Confirmation, no. (\%)} & & & & & & \\
\quad Radiology  & -- & -- & 447 (84.8) & -- & 49 (8.8) & -- \\
\quad Pathology  & -- & -- & 80 (15.2) & -- & 118 (21.1) & -- \\
\quad Cytology  & -- & -- & -- & -- & 138 (24.6) & -- \\
\quad Histopathology & 363 (100.0) & -- & -- & 712 (100.0) & 255 (45.5) & 112 (100.0) \\
\midrule
\textbf{Race, no. (\%)} & & & & & & \\
\quad White  & -- & -- & -- & 367 (100.0) & -- & -- \\
\quad Black  & -- & -- & -- & -- & -- & -- \\
\quad Asian  & -- & -- & -- & -- & -- & 112 (100.0) \\
\quad Hispanic & -- & -- & -- & -- & -- & -- \\
\quad Other  & -- & -- & -- & -- & -- & -- \\
\quad Unknown  & -- & -- & -- & -- & -- & -- \\
\midrule
\textbf{Contrast phase, no. (\%)} & & & & & & \\
\quad Portal venous  & 21 (5.8) & -- & 331 (63.5) & 367 (51.5) & 560 (100.0) & 111 (33.2) \\
\quad Arterial & 29 (8.0) & -- & 190 (36.5) & 345 (48.5) & - & 112 (33.5) \\
\quad Unknown & 313 (86.2) & -- & -- & -- & -- & 111 (33.2) \\
\midrule
\textbf{T stage, no. (\%)} & & & & & & \\
\quad I  & -- & 61 (6.1) & -- & -- & -- & 7 (6.3) \\
\quad II  & -- & 195 (19.6) & -- & -- & -- & 71 (63.4) \\
\quad III  & -- & 68 (6.8) & -- & -- & -- & 20 (17.9) \\
\quad IV & -- & 438 (44.0) & -- & -- & -- & 14 (12.5) \\
\quad Unknown & -- & -- & -- & 712 (100.0) & -- & -- \\
\midrule
\textbf{Diagnosis, no. (\%)}  & & & & & & \\
\quad PDAC  & 363 (100.0) & 996 (100.0) & 521 (100.0) & 367 (100.0) & 560 (100.0) & 112 (100.0) \\
\quad \quad $\leq 2$\,cm & 48 (13.2) & 129 (13.0) & 110 (21.1) & 30 (8.2) & 204 (36.4) & 91 (27.2) \\
\quad \quad $> 2$\,cm & 315 (86.8) & 867 (87.0) & 411 (78.9) & 337 (91.8) & 356 (63.6) & 243 (72.8) \\
\quad Non-PDAC & -- & -- & -- & -- & -- & -- \\
\quad \quad $\leq 2$\,cm & -- & -- & -- & -- & -- & -- \\
\quad \quad $> 2$\,cm & -- & -- & -- & -- & -- & -- \\
\bottomrule
\end{tabular}
}
\label{tab:external_diagnostic}
\end{table*}

\begin{table*}[t]
\centering
\caption{Characteristics of the internal (JHH) dataset used to develop and internally validate the \method\ system with a fixed train/test split.}
\resizebox{\textwidth}{!}{
\begin{tabular}{lcccc}
\toprule
\textbf{Variable} & \multicolumn{2}{c}{\textbf{Diagnostic CT scans (n = 3,440)}} & \multicolumn{2}{c}{\textbf{Normal CT scans (n = 1,645)}} \\
\cmidrule(lr){2-3} \cmidrule(lr){4-5}
 & \textbf{Training set} & \textbf{Test set} & \textbf{Training set} & \textbf{Test set} \\
\midrule
CT scan & 2,098 & 1,342 & 1,046 & 599 \\
Patient & 1,058 & 695 & 533 & 303 \\
Age, mean (SD) & 62.3 (12.7) & 66.2 (10.6) & 46.2 (12.9) & 47.7 (12.0) \\
\midrule
\textbf{Sex} & & & & \\
\quad Female, no. (\%) & 455 (43.0) & 111 (16.0) & 193 (36.2) & 109 (36.0) \\
\quad Male, no. (\%) & 505 (47.7) & 98 (14.1) & 294 (55.2) & 192 (63.4) \\
\quad Unknown, no. (\%) & 98 (9.3) & 486 (69.9) & 46 (8.6) & 2 (0.7) \\
\midrule
In-plane spacing, mm (IQR) & 0.73 (0.68, 0.79) & 0.74 (0.68, 0.80) & 0.68 (0.62, 0.73) & 0.71 (0.67, 0.77) \\
Slice thickness, mm & 0.5 & 0.5 & 0.5 & 0.5 \\
\midrule
\textbf{Race} & & & & \\
\quad White, no. (\%) & 63 (6.0) & 1 (0.1) & 381 (71.5) & 227 (74.9) \\
\quad Black, no. (\%) & 10 (0.9) & 0 (0.0) & 43 (8.1) & 51 (16.8) \\
\quad Asian, no. (\%) & 3 (0.3) & 0 (0.0) & 13 (2.4) & 13 (4.3) \\
\quad Other, no. (\%) & 3 (0.3) & 0 (0.0) & 16 (3.0) & 10 (3.3) \\
\quad Unknown, no. (\%) & 979 (92.5) & 694 (99.9) & 80 (15.0) & 2 (0.7) \\
\midrule
\textbf{Contrast phase} & & & & \\
\quad Portal venous, no. (\%) & 1,046 (49.9) & 670 (49.9) & 523 (50.0) & 299 (49.9) \\
\quad Arterial, no. (\%) & 1,052 (50.1) & 672 (50.1) & 523 (50.0) & 300 (50.1) \\
\midrule
\textbf{T stage} & & & & \\
\quad I, no. (\%) & 144 (13.6) & 26 (3.7) & -- & -- \\
\quad II, no. (\%) & 198 (18.7) & 56 (8.1) & -- & -- \\
\quad III, no. (\%) & 186 (17.6) & 33 (4.7) & -- & -- \\
\quad IV, no. (\%) & 23 (2.2) & 7 (1.0) & -- & -- \\
\quad Unknown, no. (\%) & 507 (47.9) & 573 (82.4) & -- & -- \\
\midrule
\textbf{T grade} & & & & \\
\quad I, no. (\%) & 191 (18.1) & 7 (1.0) & -- & -- \\
\quad II, no. (\%) & 288 (27.2) & 90 (12.9) & -- & -- \\
\quad III, no. (\%) & 165 (15.6) & 76 (10.9) & -- & -- \\
\quad IV, no. (\%) & 1 (0.1) & 3 (0.4) & -- & -- \\
\quad Unknown, no. (\%) & 413 (39.0) & 519 (74.7) & -- & -- \\
\midrule
\textbf{Diagnosis} & & & & \\
\quad PDAC, no. (\%) & 576 (54.4) & 288 (41.4) & -- & -- \\
\quad \quad $\leq 2$\,cm, no. (\%) & 91 (15.8) & 50 (17.4) & -- & -- \\
\quad \quad $> 2$\,cm, no. (\%) & 485 (84.2) & 238 (82.6) & -- & -- \\
\quad Non-PDAC, no. (\%) & 482 (45.6) & 407 (58.6) & -- & -- \\
\quad \quad $\leq 2$\,cm, no. (\%) & 194 (40.2) & 214 (52.6) & -- & -- \\
\quad \quad $> 2$\,cm, no. (\%) & 288 (59.8) & 193 (47.4) & -- & -- \\
\midrule
\textbf{Dilated duct} & & & & \\
\quad Yes, no. (\%) & 211 (19.9) & 263 (37.8) & 0 (0.0) & 0 (0.0) \\
\quad No, no. (\%) & 847 (80.1) & 432 (62.2) & 533 (100.0) & 303 (100.0) \\
\midrule
\textbf{Death as of Aug. 2022} & & & & \\
\quad Yes, no. (\%) & 240 (22.7) & 94 (13.5) & 1 (0.2) & 0 (0.0) \\
\quad No, no. (\%) & 720 (68.1) & 115 (16.5) & 484 (90.8) & 301 (99.3) \\
\quad Unknown, no. (\%) & 98 (9.3) & 486 (69.9) & 48 (9.0) & 2 (0.7) \\
\midrule
\textbf{Smoking history} & & & & \\
\quad Yes, no. (\%) & 34 (3.2) & 1 (0.1) & 1 (0.2) & 0 (0.0) \\
\quad No, no. (\%) & 45 (4.3) & 0 (0.0) & 32 (6.0) & 0 (0.0) \\
\quad Unknown, no. (\%) & 979 (92.5) & 694 (99.9) & 500 (93.8) & 303 (100.0) \\
\bottomrule
\end{tabular}
}
\label{tab:jhh_dataset}
\end{table*}

\begin{table*}[t]
\centering
\caption{Characteristics of external \emph{normal} CT datasets across centers.}
\resizebox{\textwidth}{!}{
\begin{tabular}{lcccccc}
\toprule
\textbf{Variable} & \textbf{E.Coast} & \textbf{N.California} & \textbf{S.California} & \textbf{C.Europe} & \textbf{N.Europe} & \textbf{E.Asia} \\
\midrule
CT scan  & 80 & 952 & 526 & 2,818 & 1,386 & 163 \\
Patient  & 80 & 952 & 526 & 2,627 & 1,386 & 105 \\
Age, mean (SD)  & 46.8 (16.7) & 63.5 (14.8) & 59.4 (18.1) & 59.2 (15.6) & 58.7 (16.5) & 60.6 (11.1) \\
\quad Unknown (no.)  & -- & -- & -- & -- & -- & -- \\
\midrule
\textbf{Sex}  & & & & & & \\
\quad Female, no. (\%)  & 27 (33.8) & 440 (46.2) & 243 (48.3) & 1,227 (46.7) & 621 (44.8) & 44 (41.9) \\
\quad Male, no. (\%)  & 53 (66.2) & 405 (42.5) & 260 (51.7) & 1,400 (53.3) & 765 (55.2) & 61 (58.1) \\
\quad Unknown (no.)  & -- & 107 (11.2) & -- & -- & -- & 0 (0.0) \\
\midrule
In-plane spacing, mm  & 0.86 (0.78, 0.94) & 0.74 (0.70, 0.82) & 0.92 (0.74, 0.91) & 0.76 (0.70, 0.84) & 0.74 (0.69, 0.78) & 0.78 (0.72, 0.85) \\
Slice thickness, mm  & 1.00 (1.00, 1.00) & 1.25 (1.25, 1.25) & 2.55 (2.00, 5.00) & 1.25 (1.00, 1.25) & 1.50 (1.50, 2.40) & 5.00 (5.00, 5.00) \\
\midrule
\textbf{Scanner, no. (\%)} & & & & & & \\
\quad TOSHIBA  & -- & -- & -- & 116 (4.1) & 517 (37.3) & -- \\
\quad SIEMENS  & -- & -- & -- & 533 (18.9) & 768 (55.4) & -- \\
\quad Philips  & -- & -- & -- & 534 (18.9) & 78 (5.6) & -- \\
\quad GE  & -- & -- & -- &  1,473 (52.3) & 21 (1.5) & -- \\
\quad Canon & -- & -- & -- & 162 (5.8) & 1 (0.1) & -- \\
\quad Unknown & -- & -- & -- & -- & 1 (0.1) & -- \\
\midrule
\textbf{Confirmation, no. (\%)} & & & & & & \\
\quad Radiology  & 80 (100.0) & 952 (100.0) & 526 (100.0) & 2,170 (82.6) & 1,169 (84.3) & 105 (100.0) \\
\quad Pathology   & -- & -- & -- & -- & 86 (6.2) & -- \\
\quad Cytology  & -- & -- & -- & -- & 40 (2.9) & -- \\
\quad Histopathology  & -- & -- & -- & 457 (17.4) & 91 (6.6) & -- \\
\midrule
\textbf{Race, no. (\%)}  & & & & & & \\
\quad White  & -- & -- & -- & 2,627 (100.0) & -- & -- \\
\quad Black  & -- & -- & -- & -- & -- & -- \\
\quad Asian  & -- & -- & -- & -- & -- & 105 (100.0) \\
\quad Hispanic  & -- & -- & -- & -- & -- & -- \\
\quad Other  & -- & -- & -- & -- & -- & -- \\
\quad Unknown  & -- & -- & -- & -- & -- & -- \\
\midrule
\textbf{Contrast phase, no. (\%)} & & & & & & \\
\quad Portal venous  & 80 (100.0) & 859 (90.2) & 327 (62.2) & 2,627 (93.2) & 1,386 (100.0) & 98 (60.1) \\
\quad Arterial  & -- & 93 (9.8) & 131 (24.9) & 191 (6.8) & -- & 65 (39.9) \\
\quad Unknown  & -- & -- & 68 (12.9) & - & -- & -- \\
\bottomrule
\end{tabular}
}
\label{tab:external_normal}
\end{table*}


\begin{table*}[t]
\centering
\caption{Reader experience in interpreting pancreatic and general abdominal CT scans for the reader study.}
\resizebox{\textwidth}{!}{
\begin{tabular}{cccccc}
\toprule
\textbf{No.} & \textbf{Reader ID} & \textbf{Experience (years)} & \textbf{CTs read per year} & \textbf{Pancreatic CTs read per year} & \textbf{Training / Expertise} \\
\midrule
1 & Specialist 1 (S1) & 37 & 600 & 200 & Pancreatic radiology \\
2 & Specialist 2 (S2) & 13 & 800 & 300 & Pancreatic radiology \\
3 & Specialist 3 (S3) & 9 & 700 & 300 & Pancreatic radiology \\
4 & General 1 (G1) & 12 & 12,000 & 1,000 & General radiology \\
5 & General 2 (G2) & 24 & 14,000 & 2,000 & General radiology \\
6 & General 3 (G3) & 7 & 8,000 & 1,000 & General radiology \\
7 & General 4 (G4) & 6 & 9,000 & 1,000 & General radiology \\
8 & General 5 (G5) & 4 & 13,000 & 1,000 & General radiology \\
9 & General 6 (G6) & 5 & 18,000 & 2,400 & General radiology \\
10 & General 7 (G7) & 5 & 10,000 & 1,200 & General radiology \\
11 & General 8 (G8) & 8 & 12,000 & 1,000 & General radiology \\
12 & General 9 (G9) & 6 & 9,000 & 1,000 & General radiology \\
13 & General 10 (G10) & 5 & 7,000 & 800 & General radiology \\
14 & General 11 (G11) & 5 & 7,000 & 600 & General radiology \\
15 & General 12 (G12) & 4 & 3,000 & 300 & General radiology \\
16 & Resident 1 (R1) & 2 & 900 & 100 & General radiology \\
17 & Resident 2 (R2) & 4 & 4,000 & 600 & General radiology \\
18 & Resident 3 (R3) & 3 & 6,000 & 800 & General radiology \\
19 & Resident 4 (R4) & 2 & 4,500 & 400 & General radiology \\
20 & Resident 5 (R5) & 2 & 2,000 & 150 & General radiology \\
21 & Resident 6 (R6) & 2 & 1,500 & 100 & General radiology \\
22 & Resident 7 (R7) & 2 & 1,600 & 100 & General radiology \\
23 & Resident 8 (R8) & 2 & 800 & 100 & General radiology \\
24 & Resident 9 (R9) & 2 & 1,000 & 100 & General radiology \\
25 & Resident 10 (R10) & 2 & 1,100 & 100 & General radiology \\
26 & Resident 11 (R11) & 2 & 700 & 100 & General radiology \\
27 & Resident 12 (R12) & 2 & 500 & 100 & General radiology \\
28 & Resident 13 (R13) & 2 & 600 & 100 & General radiology \\
29 & Resident 14 (R14) & 2 & 800 & 100 & General radiology \\
30 & Resident 15 (R15) & 2 & 1,100 & 200 & General radiology \\
\bottomrule
\end{tabular}
}
\label{tab:reader_experience}
\end{table*}

\begin{table}[t]
\centering
\caption{Overall PDAC detection performance across institutions. Sensitivity, specificity, and AUC are reported in percentage (\%). The number of evaluated patients is reported per site.}
\label{tab:pdac_performance_overall}
\begin{tabular}{llcccc}
\toprule
\textbf{cohort} & \textbf{site} & \textbf{\# of patients} & \textbf{Sensitivity} & \textbf{Specificity} & \textbf{AUC} \\
\midrule
Internal & JHH           & 581   & 97.1 & 98.7 & 98.5 \\
\midrule
\multirow{6}{*}{External}
& East Coast        & 446   & 98.9 & 96.2 & 99.4 \\
& Northern California  & 1,921 & 97.4 & 88.6 & 97.9 \\
& Southern California  & 791   & 97.7 & 89.8 & --   \\
& Central Europe      & 2,994 & 94.3 & 87.1 & 95.5 \\
& Northern Europe      & 560   & 98.9 & --   & --   \\
& Eastern Asia        & 497   & 98.8 & 92.0 & 99.4 \\
\bottomrule
\end{tabular}
\end{table}

\begin{table}[t]
\centering
\caption{PDAC detection performance across institutions. Results are reported separately for small (diameter $\leq 2$\,cm) and large (diameter $> 2$\,cm) tumors. Sensitivity and specificity are reported in percentage (\%). The number of evaluated PDAC patients is reported per site.}
\label{tab:pdac_sensitivity_specificity_by_site}

\begin{tabular}{p{0.05\linewidth}p{0.2\linewidth}P{0.2\linewidth}P{0.2\linewidth}P{0.2\linewidth}}
\toprule
\textbf{size} & \textbf{site} & \textbf{\# of patients} & \textbf{Sensitivity} & \textbf{Specificity} \\
\midrule

\multicolumn{2}{l}{\textit{Small PDAC (diameter $\leq 2$\,cm)}} & & & \\
 & Internal (JHH)                   & 43   & 95.3  & 98.7 \\
 & East Coast                       & 48   & 97.9  & 96.2 \\
 & Northern California              & 129  & 89.9  & 88.6 \\
 & Southern California              & 108  & 94.4  & 89.8 \\
 & Central Europe                   & 30   & 86.7  & 87.1 \\
 & Northern Europe                  & 204  & 98.9  & --   \\
 & Eastern Asia                     & 91   & 100.0 & 92.0 \\

\midrule

\multicolumn{2}{l}{\textit{Large PDAC (diameter $> 2$\,cm)}} & & & \\
 & Internal (JHH)                   & 236  & 97.5  & 98.7 \\
 & East Coast                       & 315  & 99.0  & 96.2 \\
 & Northern California              & 867  & 98.5  & 88.6 \\
 & Southern California              & 408  & 98.5  & 89.8 \\
 & Central Europe                   & 337  & 95.0  & 87.1 \\
 & Northern Europe                  & 356  & 98.9  & --   \\
 & Eastern Asia                     & 243  & 100.0 & 92.0 \\

\bottomrule
\end{tabular}
\end{table}

\clearpage

\end{appendices}



\end{document}